\documentclass{article}

    \usepackage[final]{project_440_550}   % keeps the footer but kills line numbers,

\usepackage[utf8]{inputenc} % allow utf-8 input
\usepackage[T1]{fontenc}    % use 8-bit T1 fonts

\usepackage[USenglish]{babel}  % there's a "canadian" option, but it's an alias for USenglish,
\usepackage{csquotes}       % smarter handling of quotes (used by biblatex)

\usepackage{booktabs}       % professional-quality tables
\usepackage{amsfonts}       % blackboard math symbols
\usepackage{nicefrac}       % compact symbols for 1/2, etc.
\usepackage{microtype}      % microtypography
\usepackage{xcolor}         % colors
\usepackage{hyperref}       % hyperlinks
\usepackage{url}            % simple URL typesetting

\usepackage[style=authoryear,maxbibnames=30]{biblatex}
\renewbibmacro{in:}{}  % drops a silly "In:" from biblatex format
\let\citet\textcite    % I do this alias because I/others find it more familiar, idk.
\let\citep\parencite   % biblatex also has a natbib option to make these,
\usepackage{amsmath}
\usepackage{graphicx}
\usepackage[capitalize,noabbrev]{cleveref}
\usepackage{subcaption}
\usepackage[shortlabels]{enumitem}
\usepackage{floatrow}

\title{Towards Robust Reinforcement Learning Market Making: Lessons from Loss Function Experimentation}

\author{%
  Siddharth Nand\\
  \texttt{sidnand@student.ubc.ca}
  \And
  \And
  Bora Guney\\
  \texttt{bguney@student.ubc.ca}
}

\begin{document}
\maketitle

\begin{abstract}
Market makers play a vital role in ensuring efficient trade within financial markets. Recent literature has delved into training market-making systems using reinforcement learning. This paper delves into the impact of various loss function formulations on risk management behavior within reinforcement-based market making. Through empirical evaluation, we contrast profit-maximizing strategies with risk-averse approaches, finding that penalties for inventory and cash management foster more resilient and adaptable market-making strategies. Our results underscore the significance of striking a balance between profit maximization and risk management in algorithmic trading, providing actionable insights for refining strategies in practical settings.
\end{abstract}

\section{Introduction}

Financial markets are essential to global economies, managing trillions in transactions daily. Market makers play a crucial role by continuously quoting buying and selling prices to ensure there always exists someone to buy and sell from. They profit by offering lower bid prices and higher ask prices, creating a spread. For example, if a market maker sets a bid price of \$10.00 and an ask price of \$10.10, buyers may purchase between \$10.05 to \$10.15, while sellers may sell between \$9.90 to \$10.20. When a buyer agrees to buy at \$10.10, the market maker sells, and when a seller agrees to sell at \$10.00, the market maker buys, earning profits from price differences and transaction volumes.

\subsection{Current Literature}

Reinforcement learning (RL) has emerged as a potent tool for optimizing market-making strategies. Its ability to discern intricate behavioral patterns from data renders it well-suited for navigating dynamic and uncertain market landscapes. Studies such as \cite{nevmyvaka2006reinforcement} and \cite{spooner2020robust} have showcased RL's potential to surpass traditional market-making models. Nonetheless, a pivotal challenge persists in formulating appropriate reward and loss functions to steer RL agents towards effective and risk-averse strategies.

Efficient market makers must adeptly manage the risks associated with inventory holdings and maintaining sufficient cash reserves. Inadequate inventory levels can lead to losses in volatile markets, while insufficient cash can curtail opportunities. \cite{guilbaud2013optimal} tackle this challenge by integrating risk considerations via constraints or implicitly shaping reward functions. However, the explicit impact of distinct loss function formulations on risk management behavior remains relatively unexplored.

In a recent study by \cite{zhang2020drlft}, the authors introduce a reinforcement learning framework tailored to market volatility dynamics, dynamically adjusting trading operations based on prevailing market conditions. Their findings underscore the efficacy of Deep Q-Network (DQN) algorithms, particularly in yielding profits despite substantial transaction costs. Noteworthy is the comparison of different RL architectures, shedding light on the underexplored potential of actor-critic models. The authors observe that actor-only models, lacking the critic component, tend to necessitate prolonged training times, suggesting a promising avenue for further exploration in market-making strategies. Additionally, \cite{zhang2020drlft} explore an actor-only model for comparison, revealing its protracted training times owing to the risk of erroneously labeling individual bad actions as good when total rewards are substantial, particularly with limited training data. Consequently, their findings suggest a promising avenue for further exploration in critic-actor models, which remain relatively under-explored in the literature.

\subsection{Problem and Contributions}

The existing literature, while valuable, leaves a crucial gap in understanding how the explicit design of loss functions shapes the balance between risk mitigation and profit maximization within RL-based market making. Our study seeks to bridge this gap, providing insights that will lead to the development of more robust and effective market-making algorithms.

We make the following key contributions:

\begin{enumerate}
    \item We enhance market-making strategies by evaluating different loss functions within a reinforcement learning framework. Our analysis reveals how incorporating penalties for excessive inventory and low cash reserves can concurrently boost profitability and mitigate risks.

    \item Through empirical evaluation, we provide valuable insights into the impact of loss function design on RL policy optimization, deepening our understanding of adaptive strategies in dynamic market environments.

    \item Our findings underscore the importance of balancing profit maximization with risk management in algorithmic trading. We offer practical implications for improved risk-aware strategies, directly benefiting the development of real-world market-making applications.
\end{enumerate}

\section{Methodology}

\subsection{Environment}

\textbf{Asset Prices}

In our simulation, we model the underlying asset's (a stock) price dynamics using geometric Brownian motion, a widely-used stochastic process in financial modeling (\cite{black1973pricing}). This process assumes asset returns are normally distributed. The governing stochastic differential equation is:

\[
dS_t = rS_t dt + \sigma S_t dW_t
\]

Here, \( r \) is the risk-free interest rate (equivalent to the savings account rate), \( \sigma \) is the volatility of the stock returns (the standard deviation of the logarithmic returns), and \( dW_t \) is a Wiener process increment representing a standard Brownian motion with mean 0 and variance \( dt \). The solution to this stochastic differential equation yields the price of the asset at time \( t \), given by:

\[
S_t = S_0 e^{\left( r - \frac{1}{2}\sigma^2 \right)t + \sigma W_t}
\]

We assume constant values for \( \sigma \) and \( r \).

\textbf{Buyer and Sellers}

Bid and ask prices are modeled as normally distributed random variables with means equal to the previous asset price and a constant standard deviation. Transactions are determined based on these prices, and profits are computed accordingly.

\subsection{Reinforcement Learning Model}

We implemented an actor-critic reinforcement learning (RL) approach, optimizing the policy function directly to maximize cumulative rewards. The actor optimizing the policy to maximize rewards and the critic estimating the expected reward through an action value function, commonly referred to as Q.

\subsection{Inputs and Outputs}

Our model relies on several key inputs for effective market-making strategies. Historical bid and ask prices, denoted by vectors $\Vec{b}$ and $\Vec{a}$ respectively, provide crucial insights into past market dynamics, each spanning a length of $t-1$ rounds of buying and selling. The profit history, encapsulated in vector $\Vec{p}$ of the same length, offers a record of accumulated profits over these rounds. Additionally, historical inventory and cash levels, represented by vectors $\Vec{i}$ and $\Vec{c}$ respectively, track inventory and cash reserves across previous rounds, aiding in decision-making. Moreover, the current inventory ($i_t$) and cash ($c_t$) values provide real-time information about available stock for selling and cash reserves for purchasing securities, shaping current trading strategies.

The output will be two real-values, a bid and an ask price at time $t+1$. This is denoted as $b_{t+1}$ and $a_{t+1}$ respectively.

\subsection{Policy Network (Actor)}

Our policy network adopts a feed-forward neural network architecture with three fully connected linear layers and ReLU activation functions. The network structure allows for adaptability to varying input sizes, enhancing flexibility in handling diverse environments.

\textbf{Loss Functions}

We experimented with two distinct loss functions and compared their performance based on total rewards:

Let \( p_t \) be profits at time \( t \), \( c_{min} \) is the threshold for acceptable cash reserves and \( i_{max} \) is the threshold for acceptable inventory levels.

\begin{enumerate}[label=\alph*)]
    \item \textbf{Profit:} This loss function solely maximizes profit, disregarding inventory and cash considerations. It is defined as:
    \[ \text{loss} = -p_t \]

    \item \textbf{Profit with Inventory and Cash Penalty:} This loss function augments the profit-oriented approach by penalizing excessive inventory and low cash reserves. It encourages the model to optimize profits while effectively managing inventory and cash flow. It is defined as:
    \[ \text{loss} = -p_t + \max(0, c_{min} - c_t) + \max(0, (i_t - i_{max}) \cdot S_t) \]
\end{enumerate}

\subsection{Incentives and Termination}

\textbf{Reward Function}

The reward function measures the performance of the agent in the environment. We use the realized gains (profit) earned from trading as the reward function.

\textbf{Termination}

Each episode will run for at-most 200 steps or until a termination condition is met. The episode terminates if the model's cash is $\leq 0$ or if inventory is $< 0$. We conducted 400 episodes in total.

\section{Results and Discussion}

\subsection{Rewards and Steps till Termination}

\begin{figure}[H]
\centering
\begin{subfigure}{0.5\textwidth}
  \centering
  \includegraphics[width=\linewidth]{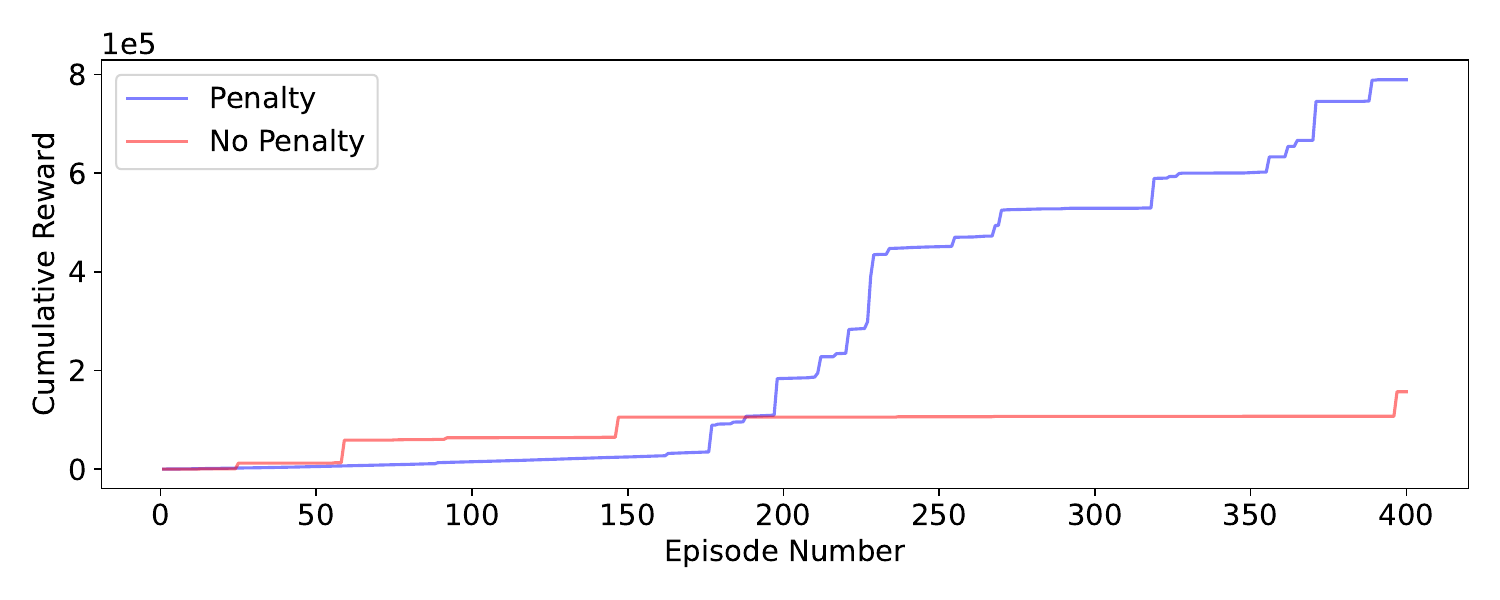}
  \caption{Cumulative Reward}
  \label{fig:cumulative_reward}
\end{subfigure}%
\begin{subfigure}{0.5\textwidth}
  \centering
  \includegraphics[width=\linewidth]{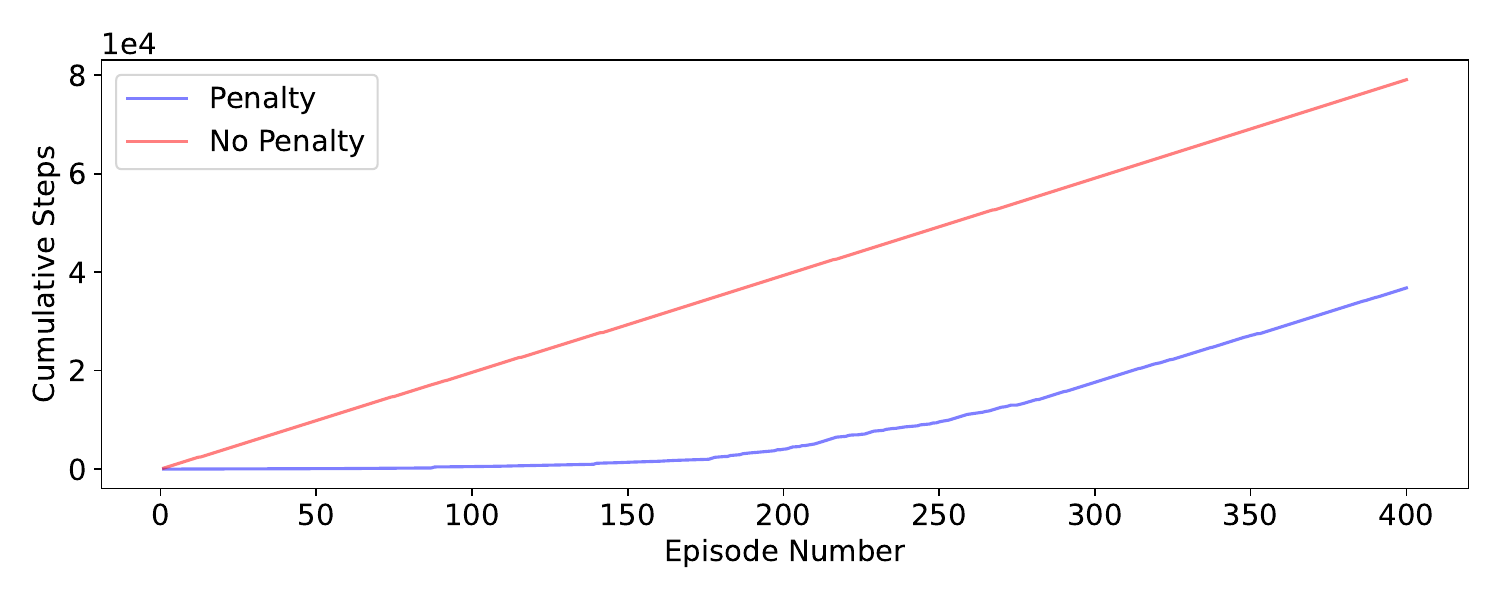}
  \caption{Cumulative Steps}
  \label{fig:cumulative_steps}
\end{subfigure}
\caption{Cumulative Reward and Steps}
\label{fig:comparison}
\end{figure}

\begin{figure}[H]
\centering
\includegraphics[width=\textwidth]{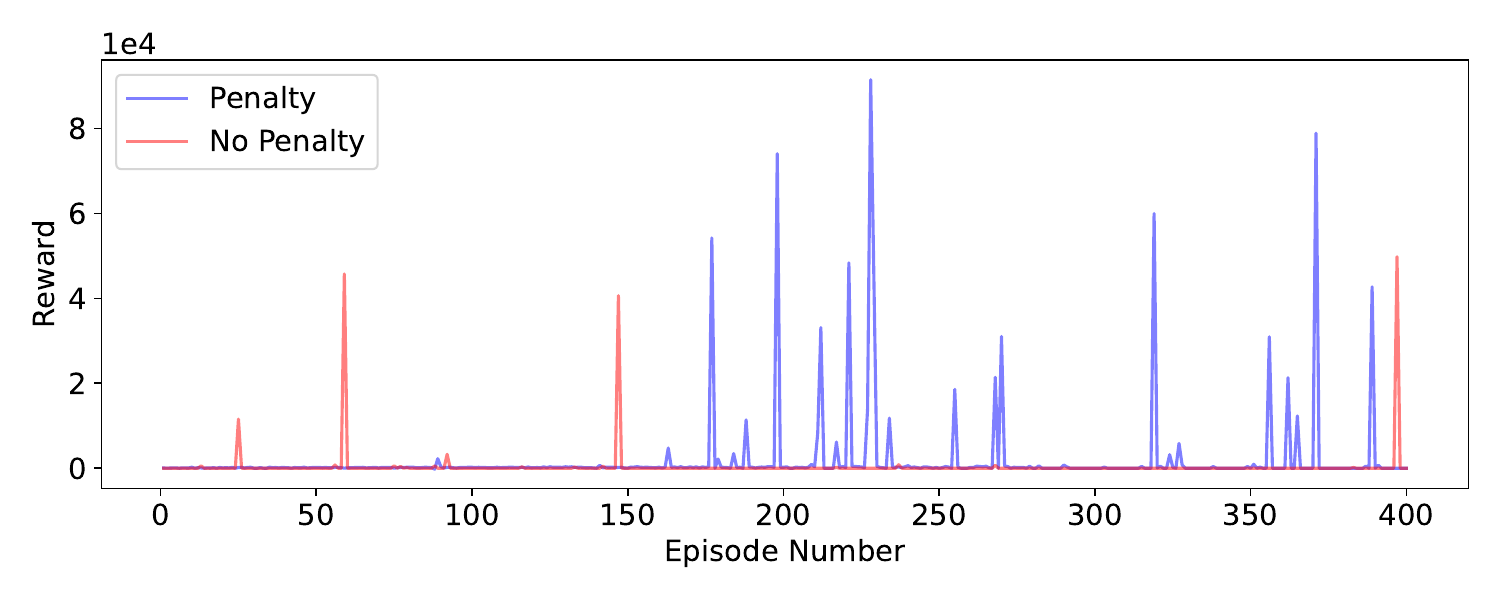}
\caption{Reward per Episode}
\label{fig:reward}
\end{figure}

Figure \ref{fig:cumulative_reward} shows the cumulative reward over the number of episodes. We can see that the agent using the penalty loss function achieves a significantly higher cumulative reward than the agent using the non-penalty loss function. This suggests that penalizing the agent for high inventory and low cash leads to better overall performance. Figure \ref{fig:cumulative_reward} for the non-penalty loss function highlights a curious phenomenon. The extended periods of stagnation, where the reward remains unchanged, suggest that the agent is minimizing its trading activity. This behavior aligns with Figure \ref{fig:reward}, where the non-penalty function repeatedly achieves a reward of 0. It's possible that the model has learned that the current market conditions make most trades unprofitable.  In response, the agent likely manipulates the bid-ask spread to discourage trading activity. This strategy allows the agent to avoid losses, even though it sacrifices potential profits.

In Figure \ref{fig:cumulative_steps}, shows the cumulative steps taken over the number of episodes. The penalty loss function initially leads to early termination, requiring fewer steps per episode. However, as the agent continues to learn, it gradually takes more steps before termination, suggesting an increased ability to navigate the environment. This suggests that the penalty-based model might initially struggle but ultimately develops strategies that allow it to sustain operations for longer durations. In contrast, the non-penalty loss function exhibits a consistent number of steps per episode, hinting at a potentially less adaptable strategy for achieving the termination condition.

In Figure \ref{fig:reward} we can see that the model with the penalty loss function achieves a consistently higher reward across episodes compared to the non-penalty model. This indicates that incorporating a penalty for high inventory and low cash into the loss function incentivizes the agent to take actions that not only generate profit but also maintain healthy inventory and cash flow levels. This strategy appears to be more successful in the long run, even though it might lead to slightly lower rewards in some individual episodes. We hypothesis this occurs because the penalty helps the agent avoid scenarios where a large profitable trade leaves it with insufficient cash for future opportunities or burdened by excess inventory that becomes difficult to sell.

\subsection{Ask-Bid Spreads}

\begin{figure}[H]
\centering
\includegraphics[width=\textwidth]{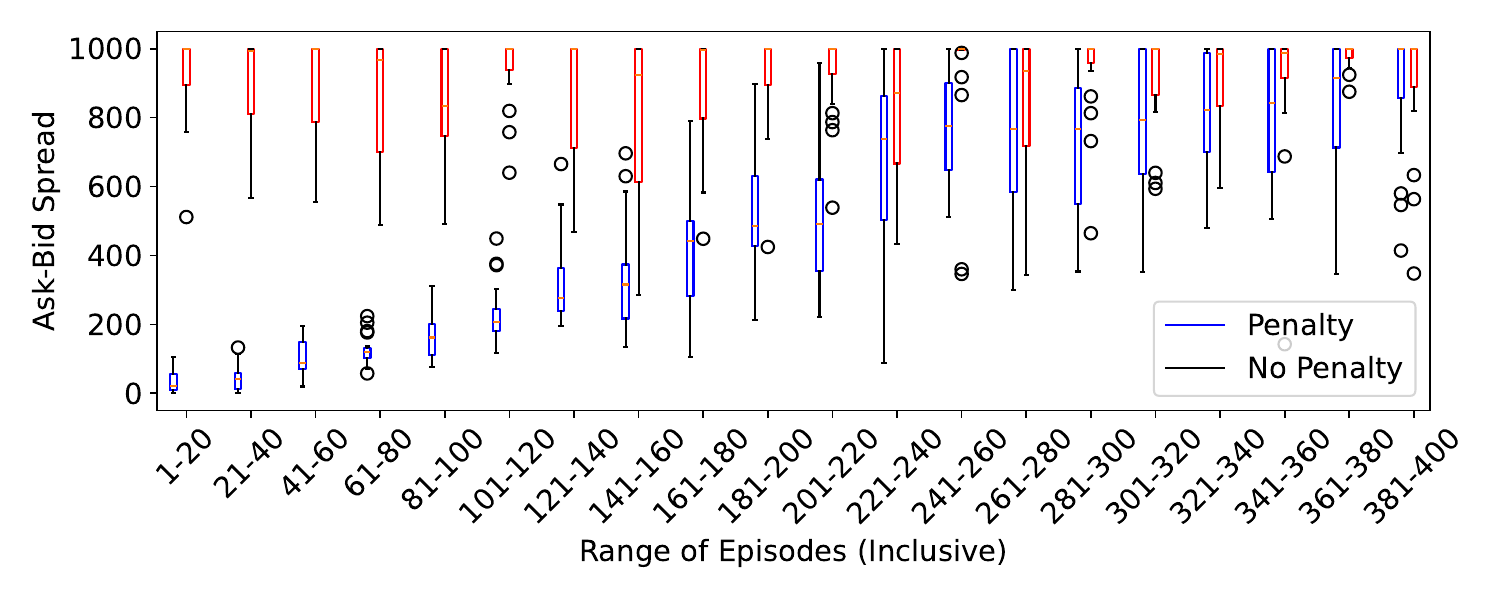}
\caption{Ask-Bid Spreads Over-Time}
\label{fig:bid-ask}
\end{figure}

Figure \ref{fig:bid-ask} reveals valuable insights into the distribution of ask-bid spreads for the two models across intervals of 20 episodes. The median spreads, represented by the lines within each box, indicate a trend where the penalty-based model generally maintains a tighter ask-bid spread compared to the non-penalty model. This suggests that the penalty-based model might be operating within a narrower profit margin, potentially focusing on higher trading frequency to compensate. Additionally, in several intervals, the penalty-based model exhibits less variability in its spread, indicated by smaller boxes (interquartile range). This hints at a more consistent strategy compared to the non-penalty model.

The presence of outliers in some intervals highlights episodes where a model's ask-bid spread deviates significantly from its normal behavior. For instance, during certain intervals, the non-penalty model exhibits noticeably wider spreads as seen by larger box sizes and more prominent outliers. This could signify that the non-penalty model has a more reactive spread strategy, possibly adjusting it drastically in response to changing market conditions or fluctuations in inventory or cash levels.

The observed ask-bid spread patterns offer clues about the different strategies and risk tolerances employed by the two models. The penalty-based model's consistently tighter spread suggests a risk-averse approach, prioritizing maintaining a healthy balance of inventory and cash flow. This strategy mitigates the risks associated with low cash reserves or excess inventory, which can hinder the agent's ability to capitalize on future opportunities. In contrast, the non-penalty model appears more opportunistic, potentially widening its spread to capture larger profits but also exposing itself to greater uncertainty.

\subsection{The Big Picture}

The analysis of the ask-bid spread, in conjunction with Figures \ref{fig:comparison} and \ref{fig:reward}, reveals a nuanced understanding of the penalty loss function's impact on the model's behavior. The penalty-based model's consistently tighter spread suggests a strategic focus on maintaining manageable inventory and cash levels, potentially sacrificing some immediate profit opportunities to ensure long-term stability. This conservative approach aligns with Figure \ref{fig:reward}, where the penalty-based model exhibits a more consistent reward across episodes, implying reduced susceptibility to market fluctuations and risky inventory imbalances. Furthermore, Figure \ref{fig:cumulative_steps} highlights how this model initially terminates episodes in fewer steps. This could signify an initial focus on quick, smaller profit trades that avoid the risks associated with holding excessive inventory or dwindling cash reserves. As the model learns a better balance, its episodes lengthen (Figure \ref{fig:cumulative_steps}), indicating that it might start to take calculated risks by holding inventory for potentially more profitable trades without compromising its overall reward (Figure \ref{fig:cumulative_reward}). This suggests a learning curve where the penalty guides the model towards a sustainable strategy.

\section{Conclusion}

Effective market making requires a delicate balance between profit maximization and risk management. Our study addresses this challenge, demonstrating how reinforcement learning models guided by tailored loss functions can develop adaptive risk-aware strategies.

\textbf{Strengths}

Our study possesses several significant strengths. First, our direct comparison of loss functions with and without explicit risk penalties provides unique insights into how these penalties shape model behavior. Results reveal that incorporating inventory and cash considerations into the loss function directly promotes risk-averse strategies, ultimately enhancing long-term profitability and stability.  Additionally, our use of an actor-critic RL architecture offers a valuable perspective within an under-explored area of market making research. Finally, our findings illuminate how loss function design influences RL policy optimization, shedding light on adaptive strategies in dynamic market environments. These integrated strengths solidify our work's distinct contributions to the field of reinforcement learning for trading.

\textbf{Weaknesses}

While our market simulation provides valuable insights, it has limitations that should be addressed in future work. Firstly, our simulation focuses on a single stock market, overlooking the diversity of global markets with distinct distributions and patterns. Employing multivariate time series simulations could mitigate this limitation by capturing the complexities of multiple market indices. Additionally, using a geometric Brown motion, which assumes normally distributed asset returns, may not fully capture market dynamics, warranting exploration of heavier-tailed distributions like the t-distribution. Moreover, our model assumes homoscedasticity in stock prices, neglecting volatility clustering inherent in financial data. Future research could incorporate heteroscedastic distributions to better represent market volatility. Furthermore, expanding our comparison of loss functions to include more sophisticated risk management metrics would enhance the robustness of our analysis.

\textbf{Future Work}

Moving forward, it is imperative for future research to delve into more sophisticated market simulations and loss functions, aiming to integrate a broader array of risk considerations into both the simulation environment and policy framework. Moreover, an important avenue for investigation lies in analyzing the behavior of actor-critic RL architectures during pivotal market scenarios, such as the onset of COVID-19 restrictions in March 2020. Understanding how these architectures adapt and evolve in response to critical market conditions is paramount for ensuring their effectiveness and resilience.

Furthermore, a promising direction for future exploration involves integrating sentiment analysis derived from financial news into actor-critic RL architectures. This is motivated by two primary factors:

\begin{itemize}
    \item There exists a level of skepticism surrounding purely quantitative trading algorithms, given that market sentiment plays a significant role in influencing stock values. By incorporating financial sentiment into actor-critic RL architectures, we can potentially enhance their risk management capabilities. This integration would enable the anticipation of future market crises within specific stocks or markets, leveraging insights from reputable financial news agencies.

    \item Despite some initial research efforts in integrating financial sentiment into RL architectures, this area remains largely underexplored. Moreover, considering the relative novelty of actor-critic RL architecture within trading algorithms, investigating this intersection presents a fertile ground for exploration. By bridging these domains, future research can unlock new avenues for enhancing trading strategies and risk management practices.
    
\end{itemize}

\newpage
\printbibliography

% \appendix

% \section{Supplementary material} \label{app:info}

\end{document}